\documentclass[conference]{IEEEtran}
\IEEEoverridecommandlockouts
\usepackage{amsmath,amssymb,amsfonts}
\usepackage{graphicx}
\usepackage[switch]{lineno}
\usepackage{adjustbox}
\usepackage{float}
\usepackage[authoryear,round]{natbib}

\usepackage{url,hyperref,microtype,subcaption}
\usepackage[onehalfspacing]{setspace}
\usepackage{cite}
\usepackage{amsmath,amssymb,amsfonts}
\usepackage{algorithmic}
\usepackage{graphicx, multirow, tabularx}
\usepackage{textcomp}
\usepackage{xcolor}
\usepackage[switch]{lineno}
\usepackage{booktabs}
\usepackage{adjustbox}
\usepackage{float}
\usepackage[utf8]{inputenc, ragged2e}
\usepackage{pifont}
\usepackage{scalerel}
\usepackage{ifthen}
\usepackage{algorithm}

\newcommand{\etal}{\emph{et al.}}

\title{FHRFormer: A Self-Supervised Masked Transformer Framework for Fetal Heart Rate Time-Series Inpainting and Forecasting}

\author{Kjersti Engan\,$^{1,*,\dagger}$, Neel Kanwal\,$^{1,\dagger}$, Anita Yeconia\,$^{2}$, Ladislaus Blacy\,$^{2}$,\\  Yuda Munyaw\,$^{1,2}$,  Estomih Mduma\,$^{2}$, Hege Ersdal\,$^{1,3}$ \vspace{0.5em} \\  
$^{1}$ University of Stavanger, Stavanger, Norway \\
$^{2}$ Haydom Lutheran Hospital, Haydom, Tanzania\\
$^{3}$ Stavanger University Hospital, Stavanger, Norway \vspace{0.5em} \\
Email: \{kjersti.engan, neel.kanwal\}@uis.no \vspace{0.5em}\\
$\dagger$ These authors contributed equally to this work.}

\begin{document}

\maketitle
\begin{abstract}
Approximately 10\% of newborns require assistance to initiate breathing at birth, and around 5\% need ventilation support. Fetal heart rate (FHR) monitoring plays a crucial role in assessing fetal well-being during prenatal care, enabling the detection of abnormal patterns and supporting timely obstetric interventions to mitigate fetal risks during labor.  Applying artificial intelligence (AI) methods to analyze large datasets of continuous FHR monitoring episodes with diverse outcomes may offer novel insights into predicting the risk of needing breathing assistance or interventions.
Recent advances in wearable FHR monitors have enabled continuous fetal monitoring without compromising maternal mobility. However, sensor displacement during maternal movement, as well as changes in fetal or maternal position, often lead to signal dropout, resulting in gaps in recorded FHR data. Such missing data limits the extraction of meaningful insights and complicates automated (AI-based) analysis.
Traditional approaches to handling missing data, such as simple interpolation techniques, often fail to preserve the spectral characteristics of the signals. 
In this paper, we propose a masked transformer-based autoencoder approach to reconstruct missing FHR signals by capturing both local temporal and frequency components of the data. The proposed method demonstrates robustness across varying durations of missing data and can be used for signal inpainting and forecasting.
The proposed approach can be applied retrospectively to research datasets to support the development of AI-based risk algorithms.  In the future, the method could be integrated into wearable FHR monitoring devices to achieve earlier and more robust real-time risk detection, although clinical benefit remains to be established.
\end{abstract}


\section{Introduction}
Despite advances in perinatal care, approximately 10\% of newborns require assistance to initiate breathing at birth, with 5\% needing advanced ventilatory support~\citep{dawes2020, wall2009neonatal}. This persistent clinical challenge highlights the indispensable role of fetal well-being assessment during labor, where continuous fetal heart rate (FHR) monitoring is a pivotal tool for detecting neonatal complications~\citep{urdal2019noise, nageotte2015fetal}. By enabling early detection of hypoxia, acidosis, and other markers of fetal distress, continuous FHR analysis empowers clinicians to intervene proactively, mitigating the risk of neonatal deaths~\citep{nageotte2015fetal, alfirevic2017}. While traditional cardiotocography (CTG) machines effectively detect fetal distress early, their high cost and need for specialized infrastructure make them impractical for many low-resource environments~\citep{mwakawanga2024barriers}. The emergence of wearable FHR monitoring devices, such as the Moyo FHR monitor~\footnote{https://shop.laerdalglobalhealth.com/}, addresses this gap in prenatal care by providing a portable, cost-effective solution that enables extended, non-invasive monitoring without substantially limiting maternal mobility 
~\citep{katebijahromi2021detection, alim2023wearable}. 
However, this increased mobility introduces other limitations, most notably signal dropouts caused by sensor displacement during maternal movement. These interruptions can result in missing data within the FHR recordings, which can compromise clinical interpretation and reduce the reliability of artificial intelligence (AI)-based analyses used to predict neonatal outcomes ~\citep{ghosh2024multi, mccoy2025intrapartum}.

\begin{figure*}[h!]
    \centering
    \includegraphics[width=0.95\textwidth]{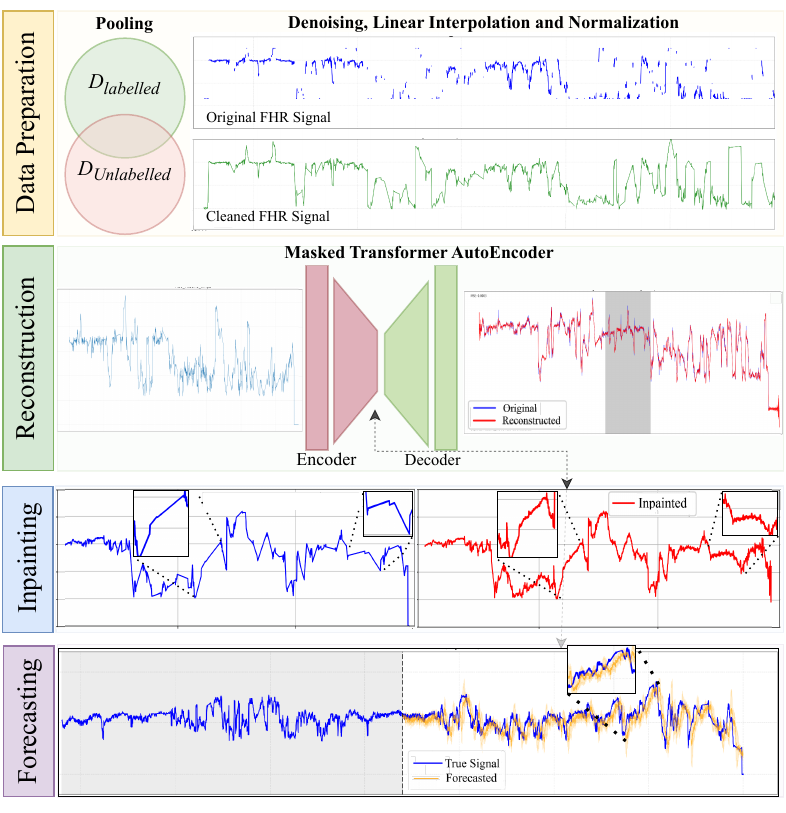}
    \caption{The application of transformer-based autoencoders in this research work: The architecture is trained on the pre-processed version of FHR data, which can perform inpainting and forecasting tasks.
    \emph{\textbf{Data preparation:}} The pooled datasets are pre-processed for Doppler noise and linearly interpolated before performing min-max normalization. \emph{\textbf{Reconstruction:}} A transformer-based encoder-decoder is trained in a self-supervised fashion to reconstruct FHR signals by masking different signal areas in each iteration.
    \emph{\textbf{Inpainting:}} Using the trained transformer-based autoencoder to reconstruct the original FHR for linearly interpolated regions to improve signal quality.
    \emph{\textbf{Forecasting:}} Feeding the context of the FHR signal to forecast iteratively for upcoming segments.}
    \label{fig:main}   
\end{figure*}

Missing data adversely impacts both time-frequency and AI-based analysis of FHR signals~\citep{barzideh2018estimation}. This is due to the potential loss of important temporal coherence and spectral features, such as short- and long-term baseline variability and deceleration patterns, that are essential for accurate risk stratification~\citep{barzideh2018estimation, rao2024automatic, cockburn2024clinical}. Therefore, addressing the signal gaps is essential to ensure the reliability of next-generation AI-enabled FHR monitoring systems~\citep{barzideh2018estimation, ghosh2024multi}. Conventional inpainting methods, such as linear or spline interpolation~\citep{spilka2012using}, ~\citep{spilka2014discriminating}, often oversimplify the physiological complexity of FHR dynamics and fail to preserve critical time-frequency relationships during prolonged signal dropouts, leading to inadequate estimation of features indicative of fetal distress. 

More advanced techniques, including wavelet-based methods~\citep{spyridou2007analysis} and sparse learning algorithms~\citep{barzideh2018estimation}, have demonstrated improved reconstruction accuracy. However, wavelet-based methods struggle with longer dropouts due to time-frequency resolution trade-offs, especially in non-stationary FHR signals, as they introduce other challenges, such as performance inefficiency for fetal state transitions.

In parallel, AI-based FHR analysis also faces fundamental limitations when relying on conventional architectures such as recurrent neural networks (RNNs) and convolutional neural networks (CNNs)~\citep{xu2023research, tahir2025bridging}. While these models are effective at detecting short-term patterns, they often struggle to capture long-range temporal dependencies that are critical for accurate FHR interpretation. Such shortcomings might compromise inpainting and forecasting performance, especially during prolonged missing data gaps.

To address these gaps, we propose a self-supervised transformer-based autoencoder framework, \emph{FHRFormer}, tailored for FHR signal reconstruction and forecasting, as illustrated in Figure~\ref{fig:main}. Trained on masked FHR data with focal-frequency loss~\citep{jiang2021focal}, FHRFormer learns both spectral and morphological representations. Unlike sequence-dependent RNNs or locality-constrained CNNs, transformers leverage multi-scale attention mechanisms~\citep{vaswani2017attention, kanwal2022attention} to globally model temporal and spectral dependencies. The frequency-aware loss enables robust reconstruction of missing segments while preserving \emph{time-domain coherence} (e.g., baseline stability, deceleration morphology) and \emph{frequency-domain characteristics} (e.g., variability power spectra). 
Integrated with the Moyo device, this framework enables the acquisition of higher-quality FHR data and introduces intelligent forecasting capabilities for potential real time early warning.

\begin{figure*}[ht!]
    \centering
    \includegraphics[width=0.95\linewidth]{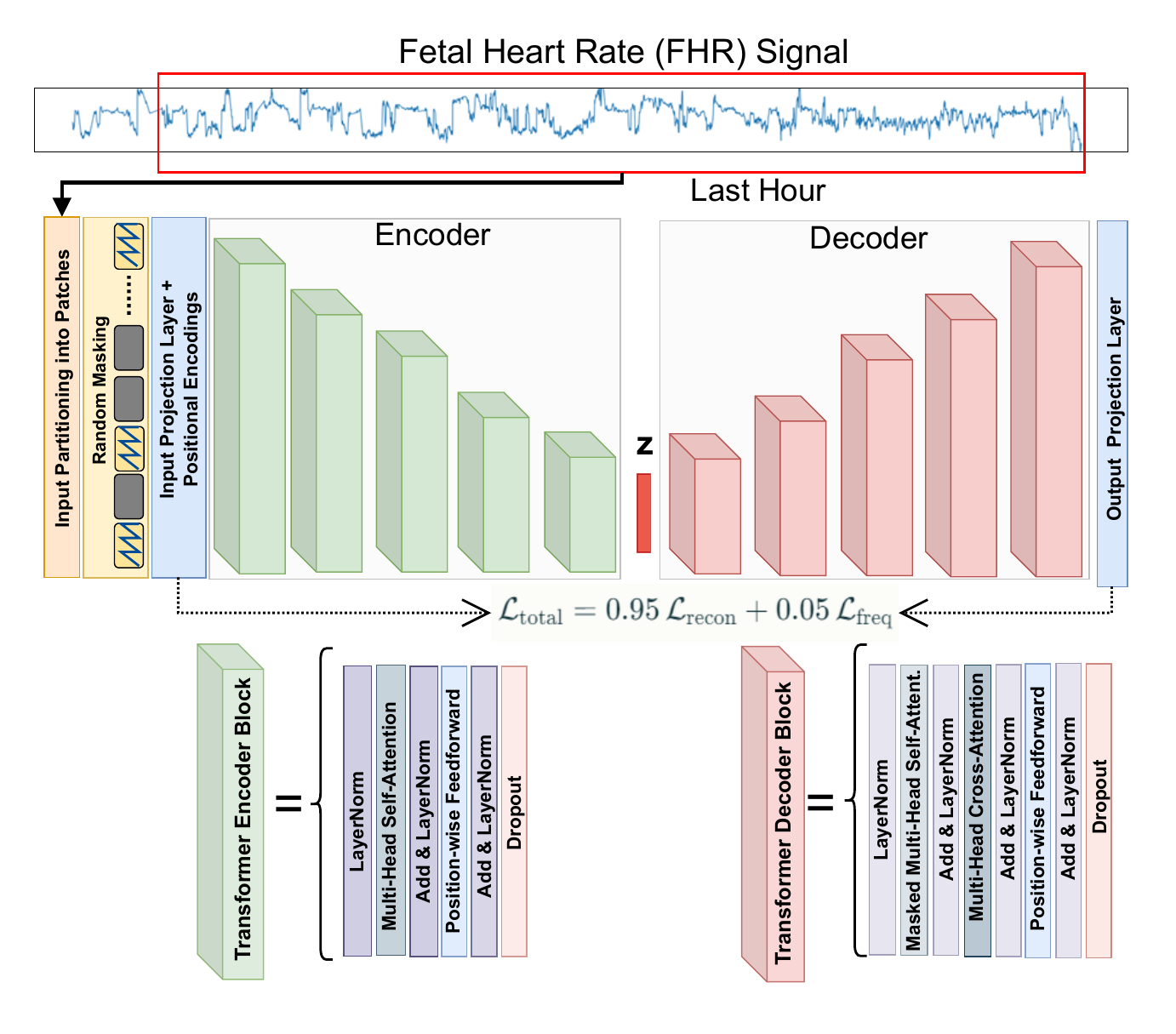}
    \caption{An overview of the 'FHRFormer' architecture. The architecture takes the fetal heart rate signal and divides it into patches. The patches are then embedded and masked before linearly projecting into the encoder. The encoder has five transformer blocks, which produce a compact representation that is then used by the decoder, which contains five transformer blocks. The final reconstruction is compared with the input using a hybrid loss.}
    \label{fig:architecture}
\end{figure*}

The long-term aims of our work are to identify during labour when intervention, such as cesarean section, is needed, and to predict which newborns will need help to start breathing at birth. Developing such prediction models requires continuous high-quality FHR recordings, and handling missing data is therefore a necessary first step.

The main contributions of this paper are: 

\begin{enumerate}
    \item A transformer-based encoder-decoder architecture for reconstructing FHR signals using synthetic masking. 
    \item A self-supervised inpainting strategy to recover missing FHR signal segments. 
    %\item A forecasting pipeline to predict FHR evolution,  enhancing utility for AI-based risk assessment. 
    \item A forecasting pipeline to predict short-term FHR evolution, demonstrating that the model captures FHR dynamics with potential utility for AI-based risk assessment.
\end{enumerate}

 %%%%%%%%%%%%%%%%%%
\subsection{Related Work}\label{sec:related}

 The problem of handling missing data in FHR monitoring has inspired the development of various techniques, ranging from classical signal processing to cutting-edge deep learning methods. To deliver a clear synthesis, we divide prior work into two categories: i) traditional signal processing and early machine learning approaches, which laid the groundwork for FHR reconstruction, and ii) modern deep learning (DL) approaches, which unlock new possibilities for modeling complex physiological signals.

% Part 1
\subsubsection{Signal Processing and Early Machine Learning Approaches}

Earlier works focusing on the missing data challenge in FHR signals relied heavily on signal processing techniques, often prioritizing simplicity over physiological fidelity. Linear interpolation~\citep{spilka2012using}, for instance, was a common choice due to its ease of implementation. However, it is a naive approach, as it does not capture the intricate variability of FHR dynamics, such as baseline fluctuations or transient decelerations~\citep{ribeiro2021non, campos2024fetal}. While more sophisticated interpolation methods (e.g., spline or cubic interpolation~\citep{cesarelli2011psd}) can better preserve signal smoothness and reduce spectral distortion compared to linear methods, they remain prone to oversmoothing transient patterns like abrupt decelerations or oscillations~\citep{campos2024fetal}. This oversimplification often makes interpolated signals unreliable, as critical spectral features are lost. In short, interpolation-based methods have low computational cost but are effective for smooth signals and smaller missing gaps. Lately, wavelet-based methods~\citep{spyridou2007analysis} emerged as a more sophisticated alternative, decomposing signals into time-frequency components to better preserve localized patterns. While effective for short gaps, wavelet-based approaches also struggled with longer dropouts, where the trade-off between time and frequency resolution became pronounced, particularly in non-stationary FHR signals driven by fetal behavioral changes~\citep{campos2024fetal}.

To overcome these limitations, sparse representation techniques emerged as a promising direction. ~\citep{oikonomou2013adaptive} proposed a two-step adaptive method for reconstructing missing FHR samples using an empirical dictionary. Their method assumed local signal stationarity, which may not hold during an active labor phase. ~\citep{da2015sparse} introduced a dictionary learning method for accurate FHR estimation using sparse decomposition of Gaussian-like functions. Classical approaches such as Gaussian Processes handle missing data effectively through probabilistic smoothing but struggle with the non-stationary, non-Gaussian nature often seen in longer FHR recordings. ~\citep{barzideh2018estimation} explored shift-invariant dictionary learning, constructing adaptive bases to reconstruct FHR signals by capturing variable shifts and transient patterns. However, their dependence on fixed dictionary sizes posed challenges when faced with variable dropout durations, which are common in real-world monitoring scenarios. In short, sparse learning methods have a higher computational cost and are prone to errors during fetal state transitions. 

The advent of machine learning in biomedical signal analysis brought new methods for handling missing FHR data, with early efforts focusing on manual feature selection. ~\citep{georgieva2013artificial} performed principal component analysis to reduce feature space and trained a feed-forward artificial neural network for classifying birth asphyxia. ~\citep{jcm7080223} performed comprehensive feature selection to train a decision tree, support vector machine, and adaptive boosting algorithm for fetal asphyxia classification. ~\citep{dash2014fetal} proposed a probabilistic method using Bayes' rule to classify FHR features. These foundational works reveal the critical need for architectures that can automatically capture hierarchical FHR patterns, paving the way for decoding FHR's physiological complexity. While these classification studies highlight the diagnostic importance of FHR features, they also emphasize the critical need for signal continuity, as missing data directly degrades the performance of such automated assessment systems.

% Part 2
\subsubsection {Deep Learning Approaches}
Recent advances in deep learning have reshaped the landscape for biomedical time-series analysis, offering tools better suited to the challenges of FHR monitoring. CNNs were widely adopted for feature extraction and classification, particularly for baseline determination and anomaly detection. For instance, ~\citep{lin2024deep} proposed the LARA (Long-term Antepartum Risk Analysis) system, which combined CNNs with information fusion operators to interpret FHR through attention visualization and deep feature analysis. Similarly, ~\citep{zhong2022ctgnet} proposed CTGNet, which utilized CNNs for FHR baseline calculation and acceleration/deceleration detection, leveraging pre-processing techniques to filter artifacts and classify signal segments. However, it is well known that CNNs often struggle with modeling long-range dependencies, limiting their ability to capture temporal patterns across extended FHR recordings~\citep{lin2024deep, zhao2019deepfhr}. Recurrent architectures, including long short-term memory (LSTM) networks, resolve this challenge of temporal patterns to some extent by leveraging their ability to model short-term dependencies in FHR signals~\citep{petrozziello2018deep}. \citep{boudet2022use} proposed FSDROP, a Gated Recurrent Unit (GRU)-based model to detect false signals in FHR, showing promising results. Nonetheless, recurrent networks can theoretically handle sequences of varying lengths; they are practically limited by vanishing gradients and the decay of hidden state information over long temporal dependencies. FHRFormer overcomes this by utilizing global self-attention, which ensures a constant path length between any two points in the FHR trace.

Transformer-based methods have gained prominence for their ability to model multi-scale dependencies through attention mechanisms~\citep{vaswani2017attention}. 
In broader medical time-series domains, like electrocardiography (ECG) and electroencephalography (EEG), transformers~\citep{zhang2022maefe,zhang2023self, wang2024convolutional} have shown promising results for various tasks by attending to both local and global patterns. However, their potential in fetal monitoring remains largely unexplored, especially for inpainting missing segments and forecasting future trends.
While self-supervised learning, particularly masked autoencoding, offers a powerful approach for leveraging unlabeled data to learn robust representations, the FHR-specific applications of self-supervised transformers are currently lacking. These architectures demonstrate the potential of deep learning for FHR analysis, yet their application to signal reconstruction remains limited by a lack of specialized temporal attention mechanisms. This absence underscores a significant need for effectively handling signal dropouts commonly encountered in FHR data from wearable devices.

Our work bridges these gaps by proposing \emph{FHRFormer}, a masked-transformer architecture tailored for FHR inpainting and forecasting. The proposed method overcomes limitations of prior traditional approaches by integrating frequency-aware loss functions~\citep{jiang2021focal} and self-supervised pre-training on unlabeled FHR data. 
This dual focus on temporal coherence and spectral fidelity positions our framework as a promising tool for FHR monitoring, particularly in addressing the challenges posed by non-stationary FHR dynamics.

\section{Materials and Methods}  \label{sec:methodology}

This section presents details on the FHR data used for training and validation, describes data preprocessing for pre-training, the structure of the transformer-based autoencoder, fine-tuning for the forecasting task, the experimental setup, and the evaluation metrics used for performance evaluation. 

\subsection{Data Material}
The FHR data were obtained through the Safer Births project~\footnote{saferbirths.com}, a collaborative research initiative involving international research institutions and hospitals in Tanzania. Data collection occurred in two phases at two urban and one rural Tanzanian hospital between October 2015 and June 2018, and again between December 2019 and July 2021, encompassing a total of 5,225 recorded labors from 5,225 distinct participants (one recording per participant). This study focuses exclusively on labors initially assessed as normal upon hospital admission.
 The data were acquired using the Laerdal Moyo FHR monitor~\citep{laerdal_moyo}. This device comprises a primary unit displaying the measured heart rate to healthcare personnel and a sensor unit integrating a Doppler ultrasound sensor and an accelerometer. The sensor unit is secured to the mother using an elastic strap. The device is programmed to alert healthcare personnel if the detected FHR remains outside the range of 110-160 beats per minute for 10 minutes or outside the range of 100-180 beats per minute for 3 minutes. The FHR is measured via a $1\,\mathrm{MHz}$, $5\,\mathrm{mW/cm^2}$ pulsed wave Doppler ultrasound sensor, sampled at $2\,\mathrm{Hz}$, generating a discrete FHR signal, $\mathbf{x} \in \mathbb{R}^{L \times 1}$,  where $L$ is the FHR signal length. 
 
\subsection{Pre-processing}

Doppler shift error can result in a doubling or halving of the perceived frequency, and thereby the estimated heart rate.  
Additionally, periods of missing data may occur due to sensor displacement caused by maternal movement, as previously discussed. 
To mitigate these issues, we adopted the noise reduction algorithm proposed by Urdal \etal~\citep{urdal2019noise}, selected because it was developed and validated specifically for Doppler-based FHR signals recorded with the Moyo monitor within the same Safer Births cohort, making its noise model directly applicable to our data without requiring re-validation on a different acquisition system.
This method addresses frequency-doubling/halving artifacts and applies linear interpolation as an initial step to fill missing signal segments. A binary mask indicating the locations of missing data was retained to enable refined inpainting using the proposed FHRFormer framework.

All FHR signals were trimmed or zero-padded to a fixed length of 7,200 time steps, corresponding to one hour of physiological time-series data. The FHR signals were normalized to [0,1], where 0 corresponds to 0 bpm, and 1 corresponds to the global maximum of 220 bpm. 
We focus on the final hour preceding birth to provide FHRFormer with sufficient temporal context (7,200 samples) to model long-range variability trends. This is also the window during which intervention decisions are most crucial, and FHR patterns are most predictive of immediate neonatal outcome.

The pre-processed FHR data set, $\mathcal{X}=\{{\mathbf x}^{(j)}\}$, where $j$ is the episode index, often omitted for simplicity, was divided into three non-overlapping subsets at the episode (patient) level into training set $\mathcal{X}_{tr}$, validation set $\mathcal{X}_{val}$, and test set $\mathcal{X}_{test}$, consisting of 4,486, 369, and 370 FHR episodes, respectively. Participants were assigned to subsets via a single random shuffle drawn across the full pooled dataset, spanning both collection phases (2015–2018, 2019–2021) and all three hospital sites, rather than by chronological order; consequently, no subset is concentrated within a narrow time window or a single site.
At the patch level ($p_s=30$), each one-hour recording yields $N=240$ non-overlapping patches, so the training set alone provides over one million (4,486 $\times$ 240 $\approx$ 1.08 million) individual patch instances for the self-supervised reconstruction objective. This is comparable to other single-institution biomedical masked-autoencoder studies \citep{zhang2022maefe,zhang2023self} and adequate for training the five-layer transformer autoencoder from random initialization.
Because assignment was performed at the level of complete recordings rather than by time segment, no signal segment used for training, validation, or testing originates from the same labor episode as another, and adjacent segments within a single continuous recording are never split across subsets. This also promotes a higher fraction of training data to augment generalizability.

\subsection{FHRFormer Architecture}
\label{sec:FHRFormer}
Figure~\ref{fig:architecture} shows a detailed overview of the proposed FHRFormer architecture along with the training strategy. The following sections further describe the components of the architecture.

\subsubsection{Input Patchification}
Given a univariate FHR time-series signal $\mathbf{x} \in \mathbb{R}^{L \times d}$, where $L$ is the FHR length and $d$ is the signal dimensionality (here $d=1$). The input is first divided into $N=L/p_s$ non-overlapping segments, or patches of length $p_s$.

\begin{equation}
\mathbf{x} = [\mathbf{x}_1, \mathbf{x}_2, \ldots, \mathbf{x}_N], \quad \mathbf{x}_i \in \mathbb{R}^{p_s \times d}
\end{equation}

The purpose of patching is to allow the transformer model to process manageable local segments while still retaining the ability to learn dependencies across the entire time sequence. We have experimented with multiple patch sizes, $p_s =(30, 60, 120, 240, 480$).

\subsubsection{Randomized Masking Strategy}
The masking strategy facilitates robust feature learning in a self-supervised manner by forcing the model to infer and reconstruct missing physiological information from the available surrounding context. Masking is applied randomly at the patch level using a binary mask vector $\mathbf{m} \in \{0, 1\}^N$, where $m_i = 0$ indicates that patch $i$ is masked. We define the following sets:
$\mathcal{M} = \{i \mid m_i = 0\}$ as the indices of masked patches, and $\mathcal{U} = \{i \mid m_i = 1\}$ as the indices of unmasked (visible) patches. To maintain a consistent learning signal, at least one patch is always masked. 
The parameter $\gamma \in$ (0, 100] denotes the target masking ratio, expressed as a percentage (as reported in Table~\ref{tab:implementation} and the Results), such that the number of masked patches per signal is $|M|$ = max(1, round(($\gamma$/100) $\cdot$ $N$)), with at least one patch always masked regardless of $\gamma$ or $N$.

\subsubsection{Input Embedding and Positional Encoding}
Each FHR patch $\mathbf{x}_i$ is flattened and projected into a higher-dimensional latent space using a learnable linear projection:

\begin{equation}
\mathbf{e}_i = \mathbf{W}_{in} \cdot \mathrm{vec}(\mathbf{x}_i) + \mathbf{b}_{in}
\end{equation}

where $\mathbf{W}_{in} \in \mathbb{R}^{d_{\text{model}} \times (p_s \cdot d)}$ and $\mathbf{b}_{in}$ represent the learnable weights and bias, respectively. The latent dimension $d_{\text{model}}$ allows the transformer to represent local signal morphologies in a rich, learnable space. To preserve the temporal order of the sequence, sinusoidal positional encodings $\mathbf{p}_i$ are added to the embeddings. The input to the encoder consists only of the visible patches:

\begin{equation} \label{eq:3}
\tilde{\mathbf{e}}_i = \mathbf{e}_i + \mathbf{p}_i, \quad \forall \quad i \in \mathcal{U}
\end{equation}

The mask token (i.e., a shared learnable embedding) $\mathbf{e}_{mask} \in \mathbb{R}^{d{\text{model}}}$ serves as a placeholder for the missing physiological segments during the reconstruction process. For the decoder, 
$\mathbf{e}_{mask} + \mathbf{p}_i$ is used at all masked positions $i \in \mathcal{M}$.

\subsubsection{Transformer Encoder}
The encoder consists of a stack of $L=5$ identical blocks. Each block comprises: i) Multi-head Self-attention (MHSA), allowing visible patches to attend to one another globally; ii) a Position-wise Feed-forward Network (FFN); and iii) residual connections, with Layer Normalization (LN) applied before each sub-layer (Pre-Norm). The Position-wise FFN consists of two linear layers with a GeLU activation, applied independently to each patch to enhance non-linear feature transformation. We adopt the "Pre-Norm" formulation for improved training stability, where normalization precedes both the MHSA and FFN sub-layers. Let $\mathbf{h}_i^{(0)} = \tilde{\mathbf{e}}_i$ be the input to the first encoder layer. For each layer $\ell \in \{1, \dots, L\}$, the state updates are defined as:

\begin{equation}
\begin{aligned} \label{eq:4}
\mathbf{z}_i^{(\ell)} &= \text{MHSA}(\text{LN}(\mathbf{h}_i^{(\ell-1)})) + \mathbf{h}_i^{(\ell-1)} \\
\mathbf{h}_i^{(\ell)} &= \text{FFN}(\text{LN}(\mathbf{z}_i^{(\ell)})) + \mathbf{z}_i^{(\ell)}
\end{aligned}
\end{equation}

where $\mathbf{z}_i^{(\ell)}$ is the intermediate latent representation. Dropout is applied after each sub-block to discourage co-adaptation of neurons and promote redundant feature learning. In the MHSA mechanism, the attention matrix is fully connected and lacks a causal mask (i.e., no look-ahead restriction). This allows the encoder to leverage both preceding and succeeding signal context to reconstruct a missing segment. The final output of the encoder for visible patches is denoted as $\mathbf{Z} = \{ \mathbf{h}_i^{(L)} \mid i \in \mathcal{U} \}$.

\subsubsection{Transformer Decoder}
The decoder reconstructs the full FHR signal by processing both the visible encoder outputs and the learnable mask tokens. The input to the first decoder layer, $\mathbf{d}_i^{(0)}$, is constructed as:

\begin{equation} \mathbf{d}_i^{(0)} = \begin{cases} \mathbf{h}_i^{(L)}, & i \in \mathcal{U} \\ \mathbf{e}_{mask} + \mathbf{p}_i, & i \in \mathcal{M} \end{cases}\label{eq:5} \end{equation}

This forms the complete sequence $\mathbf{D}^{(0)} = [\mathbf{d}_1^{(0)}, \dots, \mathbf{d}_N^{(0)}]$. Each decoder block follows a Pre-Norm Transformer formulation and consists of three sub-layers:  Multi-Head Self-Attention (MHSA), Multi-Head Cross-Attention (MHCA) over the encoder outputs, and a position-wise Feed-Forward Network, each wrapped with residual connections. 
Given the non-causal nature of signal inpainting, bidirectional attention is employed throughout to enable global reasoning over the temporal context. 

The updates for each decoder layer $\ell \in {1,\ldots L}$ are defined as:

\begin{equation}
\begin{aligned}
\tilde{\mathbf{d}}^{(\ell)}_i &= \mathbf{d}^{(\ell-1)}_i + \text{MHSA}\big(\text{LN}(\mathbf{d}^{(\ell-1)}_i)\big), \\
\hat{\mathbf{d}}^{(\ell)}_i &= \tilde{\mathbf{d}}^{(\ell)}_i + \text{MHCA}\big(\text{LN}(\tilde{\mathbf{d}}^{(\ell)}_i), \mathbf{Z}\big), \\
\mathbf{d}^{(\ell)}_i &= \hat{\mathbf{d}}^{(\ell)}_i + \text{FFN}\big(\text{LN}(\hat{\mathbf{d}}^{(\ell)}_i)\big).
\end{aligned}
\end{equation}
The output of the final decoder layer is denoted as $\mathbf{D} = [\mathbf{d}^{(L)}_1, \dots, \mathbf{d}^{(L)}_N] \in \mathbb{R}^{N \times d_{\text{model}}}$, representing contextualized embeddings for all patches including both masked and unmasked positions.

To reconstruct the original FHR signal, a linear output projection layer maps these high-dimensional embeddings back to the patch space:
\begin{equation}\label{eq:7}
    \hat{\mathbf{x}}_i = \mathbf{W}_{\text{out}} \mathbf{d}_i + \mathbf{b}_{\text{out}}, \quad \text{where} \quad \mathbf{W}_{\text{out}} \in \mathbb{R}^{(p_s \cdot d) \times d_{\text{model}}}
\end{equation}
Here, $p_s$ is the patch size and $d$ is the feature dimension of the original signal; thus $\hat{\mathbf{x}}_i \in \mathbb{R}^{p_s \times d}$ represents the reconstructed $i$-th patch.
 
Normally, only the masked part is reconstructed, and the unmasked part is kept as the original, giving the reconstructed signal as $X^{R} =[x_{0}^{R} \ldots x_{N}^{R}]$ with:  
\begin{equation}
%\tilde{\mathbf{z}}_i =
\mathbf{x}_{i}^{R} =
\begin{cases}
\mathbf{x}_i & i \in \mathcal{U} \\
\hat{\mathbf{x}}_{i} , & i \in \mathcal{M}
\end{cases}
\end{equation}

\subsubsection{Reconstruction Objective} 
\label{sec:objective}
The objective of FHRFormer is to reconstruct masked regions of the FHR signal as closely as possible. Let $\mathcal{M}$ be the set of indices corresponding to withheld (masked) patches. The \textit{primary loss} is the mean squared error on the masked patches.

\begin{equation} \label{eq:rec}
    \mathcal{L}_{\text{recon}} = \frac{1}{|\mathcal{M}|} \sum_{i \in \mathcal{M}} \|\mathbf{\hat{x}}_i - \mathbf{x}_i\|^2
\end{equation}

To further guide the model toward preserving essential spectral (frequency) physiological characteristics of FHR signals, a frequency-domain loss, namely the focal frequency loss~\citep{jiang2021focal}, %is combined with the reconstruction loss to develop a hybrid loss function.
is incorporated to emphasize discrepancies in the frequency content of the reconstructed signal. 
For each masked patch of length $p_s$, the Discrete Fourier Transform (DFT) magnitude spectrum is computed using a Hann window. The per-patch frequency-domain loss is defined as:

\begin{equation} \label{eq:freq}
l_{\text{f}}^{i} = \frac{1}{K} \sum_{k=1}^{K} \left(1 - e^{-\left||\mathcal{F}(\mathbf{x_{i}})_k| - |\mathcal{F}(\mathbf{\hat{x}_i})_k|\right|}\right)^\beta \cdot \left||\mathcal{F}(\mathbf{\hat{x}_i})_k| - |\mathcal{F}(\mathbf{x_i})_k|\right|
\end{equation}

The overall frequency-domain loss is obtained by averaging over all masked patches:

\begin{equation} \label{eq:freq}
\mathcal{L}_{\text{freq}} = \frac{1}{|\mathcal{M}|}\sum_{i \in \mathcal{M}} l_{\text{f}}^{i} 
\end{equation}

Here $|\mathcal{F}(\mathbf{x_i})_k|$ and $|\mathcal{F}(\mathbf{\hat{x}_i})_k|$ denote the amplitude (magnitude) of the $k$-th Fourier transform coefficient for the target signal $\mathbf{x_i}$ and the predicted signal $\mathbf{\hat{x}_i}$, respectively, and $K$ is the number of frequency bins.  The parameter $\beta$ (fixed to 1) controls the weighting of the loss based on the discrepancy between the target and predicted amplitudes.

The final training objective is a weighted combination of the reconstruction and frequency-domain losses:

\begin{equation} \label{eq:tot}
    \mathcal{L}_{\text{total}} = \alpha \cdot \mathcal{L}_{\text{recon}} + (1 - \alpha) \cdot \mathcal{L}_{\text{freq}}
\end{equation}

where $\alpha \in[{0,1}]$ balances pointwise accuracy and spectral fidelity, we use $\alpha = 0.95$. The inclusion of the frequency-domain loss encourages reconstructions that not only minimize pointwise errors but also preserve the underlying periodicity and spectral structure critical for physiological interpretation.
Algorithm~\ref{alg:MTA} summarizes the overall training procedure for \emph{FHRFormer}. 

\begin{algorithm}[h]
\caption{Self-supervised Training of FHRFormer} \label{alg:FHRFormer}
\begin{algorithmic}[1]
\STATE \textbf{Input:} Training set $\mathcal{X}_{tr}$, Validation set $\mathcal{X}_{val}$, masking ratio $\gamma$, patch size $p_s$, initial parameters $\theta$, learning rate $\eta$, max epochs $E$, patience $Pat$
\STATE Initialize optimizer with $\eta$; $\mathcal{L}_{best} \leftarrow \infty$; $c \leftarrow 0$
\FOR{epoch $= 1$ to $E$}
\STATE Set model to \texttt{training} mode
\FOR{each batch $\mathbf{x}_{b}$ in $\mathcal{X}_{tr}$}
\STATE $\mathbf{x}_{norm} \leftarrow \text{MinMaxNormalize}(\mathbf{x}_b)$
\STATE Partition $\mathbf{x}_{norm}$ into $N$ patches $\{\mathbf{x}_i\}_{i=1}^N$ of size $p_s$
\STATE Generate random binary mask $\mathbf{m}$ with ratio $\gamma$; define $\mathcal{U}$ and $\mathcal{M}$
\STATE $\tilde{\mathbf{e}}_i \leftarrow \text{Embed}(\mathbf{x}_i) + \mathbf{p}_i, \quad \forall \quad i \in \mathcal{U}$ // See Eq.~\eqref{eq:3}
\STATE $\mathbf{Z} \leftarrow \text{Encoder}(\{\tilde{\mathbf{e}}_i\}_{i \in \mathcal{U}})$ 
// Latent features via Eq.~\eqref{eq:4}
\STATE Assemble full sequence using Eq.~\eqref{eq:5}
\STATE $\hat{\mathbf{x}} \leftarrow \text{Decoder}(\mathbf{D}^{(0)}, \mathbf{Z})$ 
// Global reconstruction via Eq.~\eqref{eq:7}
\STATE $\mathcal{L}_{\mathrm{total}} \leftarrow \alpha \mathcal{L}_{\mathrm{MSE}}(\mathbf{x}_{i \in \mathcal{M}}, \hat{\mathbf{x}}_{i \in \mathcal{M}}) + (1-\alpha) \mathcal{L}_{\mathrm{freq}}$ // Evaluate only on $\mathcal{M}$
\STATE Update $\theta$ via backpropagation and optimizer
\ENDFOR
\STATE Set model to \texttt{evaluation} mode
\STATE Compute $\mathcal{L}_{\mathrm{val}}$ on $\mathcal{X}_{val}$ (unseen patient split)
\IF{$\mathcal{L}_{\mathrm{val}} < \mathcal{L}_{best}$}
\STATE $\mathcal{L}_{best} \leftarrow \mathcal{L}_{\mathrm{val}}$; Save $\theta_{best} \leftarrow \theta$; $c \leftarrow 0$
\ELSE
\STATE $c \leftarrow c + 1$ // counter to patience
\ENDIF
\IF{$c \geq Pat$}
\STATE \textbf{Break} // Early stopping to prevent overfitting
\ENDIF
\ENDFOR
\STATE \textbf{Output:} Optimized model parameters $\theta_{best}$
\end{algorithmic} \label{alg:MTA}
\end{algorithm}

\subsection{Implementation Details}

We used Python and PyTorch for scripting and architecture design. As the architecture was custom-designed, no pretrained weights were available, so all parameters were initialized randomly. Hyperparameters were explored through a grid search for improved validation metrics. The final chosen parameters were the Adam optimizer with a weight decay of 0.01, the \emph{ReduceLROnPlateau} scheduler with a learning rate of 0.0001, early stopping of 20 epochs over the validation loss to avoid overfitting, the batch size of 128, and a fixed random seed for reproducibility. 
%Finally, we used the best-performing model weights to report evaluation metrics on the validation and hold-out sets. 
The model selection was based exclusively on the lowest total loss ($\mathcal{L}$, as shown in Table~\ref{tab:implementation}) observed on the validation set $\mathcal{X}_{val}$ during training, following the early-stopping criterion in Algorithm~\ref{alg:FHRFormer}. The corresponding checkpoint $\theta_{best}$ was frozen and used, without further tuning, to compute all metrics reported on both $\mathcal{X}_{val}$ and the independent hold-out test set $\mathcal{X}_{test}$, which was not accessed at any point during model or hyperparameter selection.
All chosen hyperparameters for the FHRFormer architecture and training are summarized in Table~\ref{tab:implementation}. All training and inference experiments were performed on a Tesla V100-PCIE GPU with 32 GB of memory.
The source code and model weights are available on~\href{https://github.com/NeelKanwal/FHRFormer-A-Self-supervised-Transformer-Approach-for-Fetal-Heart-Rate-Inpainting-and-Forecasting}{GitHub}.

\begin{table*}[ht!]
    \centering
    \caption{The configuration and hyperparameters used for the training of the proposed FHRFormer model.}
     \resizebox{0.8\textwidth}{!}{
    \begin{tabular}{||l| l||}
    \hline
       {\textbf{\vspace{0.1em}Parameter}} & \textbf{Value} \\ 
       \hline
        Number of encoder layers & 5 \\
        Number of decoder layers & 5 \\
        Number of attention heads & 16 \\
        Dimension of intermediate layer & 1024 \\ 
        Features in decoder input ($d_{model}$) & 512 \\
        Input patch size ($p_s$) & \{30, 60, 120, 240, 480\} \\
        Total length of input ($L$) & 7200 \\
        Masking ratio ($\gamma$ \%) & \{5, 10, 15, 20, 25, 30, 35\} \\
        Input normalization & min-max \\
        Early stopping (patience $Pat$) & 20 epochs\\ 
        Batch size & 128 \\
        Scheduler & ReduceLROnPlateau with patience=5\\
        Optimizer & Adam \\
        Learning rate ($\eta$) & 0.0001 \\
        Dropout regularization & 0.1 \\
        Loss function ($\mathcal{L}$) & 0.95$\times$($\mathcal{L}_{recon}$) + 0.05$\times$ ($\mathcal{L}_{freq}$)\\
        \hline
    \end{tabular}}
    \label{tab:implementation}
\end{table*}

\subsection{Evaluation Metrics}
For comprehensive performance assessment of FHRFormer, we report the following metrics: Reconstruction Loss, Peak Signal-to-Noise Ratio (PSNR), Structural Similarity Index Measure (SSIM), Fréchet Inception Distance (FID), Mean Squared Error (MSE), Root Mean Squared Error (RMSE), Mean Absolute Error (MAE), and Correlation Coefficient (CC). 

\textit{Reconstruction Loss (RL)} quantifies the difference between the original and reconstructed FHR, as defined by the loss function in~\eqref{eq:rec}. \textit{Mean Squared Error (MSE)} is calculated as the average squared difference between the target and reconstructed signals, given by $ \mathrm{MSE} = \frac{1}{N} \sum_{i=1}^N (\mathbf{x}_i - \hat{\mathbf{x}}_i)^2 $, where $\mathbf{x}_i$ and $\hat{\mathbf{x}}_i$ denote original and reconstructed values, respectively.

\begin{table*}[ht!]
    \centering
    \caption{The validation and test results. The performance of the transformer-based autoencoder for different input patch sizes. The best results are highlighted in bold, and the second-best are underlined for each subset.\vspace{0.5em}}
    \renewcommand{\arraystretch}{1.9} % Adjust row height for better readability
    \begin{tabularx}{\textwidth}{|>{\Centering\arraybackslash}m{0.5cm} |>{\Centering\arraybackslash}m{1.3cm} |>{\Centering\arraybackslash}m{1.5cm} |>{\Centering\arraybackslash}m{1.5cm} |>{\Centering\arraybackslash}m{1.5cm} |>{\Centering\arraybackslash}m{1.5cm} |>{\Centering\arraybackslash}m{1.5cm} |>{\Centering\arraybackslash}m{1.5cm} |>{\Centering\arraybackslash}m{1.5cm} |>{\Centering\arraybackslash}m{1.5cm} |}
       \cline{1-10}
       \hline\hline
         {} & \textbf{\shortstack{Input\\ Patch\\ Size ($p_s$)}} & \textbf{\shortstack{RL ($\downarrow$)}} & \textbf{\shortstack{PSNR ($\uparrow$)}} & \textbf{\shortstack{SSIM ($\uparrow$)}} & \textbf{\shortstack{FID ($\downarrow$)}} & \textbf{\shortstack{MSE ($\downarrow$)}} & \textbf{\shortstack{RMSE ($\downarrow$)}} & \textbf{\shortstack{MAE ($\downarrow$)}} & \textbf{\shortstack{CC ($\uparrow$)}} \\ 
    \cline{1-10}
    \hline\hline
        \multirow{5}{*}{\rotatebox[origin=c]{90}{\textbf{Validation}}} 
        & 30  & \textbf{0.051}  & \textbf{41.55} & \textbf{0.9981} & \textbf{0.503} & \textbf{0.000071} & \textbf{0.0084} & \textbf{0.0041} & \textbf{0.9992} \\ 
        \cline{2-10}
        & 60  & \underline{0.205}  & \underline{35.63} & \underline{0.9925} & \underline{1.965} & \underline{0.000273} & 0.1653 & \underline{0.0088} & \underline{0.9971} \\ 
        \cline{2-10}
        & 120 & 0.489  & 31.58 & 0.9822 & 4.985 & 0.000694 & \underline{0.0264} & 0.0149 & 0.9925 \\ 
        \cline{2-10}
        & 240 & 1.264 & 27.72 & 0.9615 & 12.051 & 0.001695 & 0.0411 & 0.0244 & 0.9816 \\ 
        \cline{2-10}
        & 480 & 3.232 & 23.01 & 0.9092 & 34.242 & 0.004994 & 0.0707 & 0.0433 & 0.9454 \\ 
        \cline{1-10}
       \hline\hline
        \multirow{5}{*}{\rotatebox[origin=c]{90}{\textbf{Test}}} 
        & 30  & \textbf{0.048}  & \textbf{41.38} & \textbf{0.9984} & \textbf{0.522} & \textbf{0.000072} & \textbf{0.0085} & \textbf{0.0040} & \textbf{0.9993} \\ 
        \cline{2-10}
        & 60  & \underline{0.204}  & \underline{35.56} & \underline{0.9941} & \underline{1.999} & \underline{0.000278} & \underline{0.0166} & \underline{0.0086} & \underline{0.9976} \\ 
        \cline{2-10}
        & 120 & 0.537  & 31.57 & 0.9862 & 4.987 & 0.000952 & 0.02637 & 0.0145 & 0.9942 \\ 
        \cline{2-10}
        & 240 & 1.228 & 27.71 & 0.9703 & 12.041 & 0.001692 & 0.0411 & 0.0240 & 0.9854 \\ 
        \cline{2-10}
        & 480 & 3.262 & 22.95 & 0.9275 & 34.663 & 0.005061 & 0.0711& 0.0435 & 0.9553 \\ 
        \cline{1-10}
        \hline\hline
    \end{tabularx}
    \label{tab1}
\end{table*}

\begin{figure*}[h!]
    \includegraphics[width=0.95\linewidth]{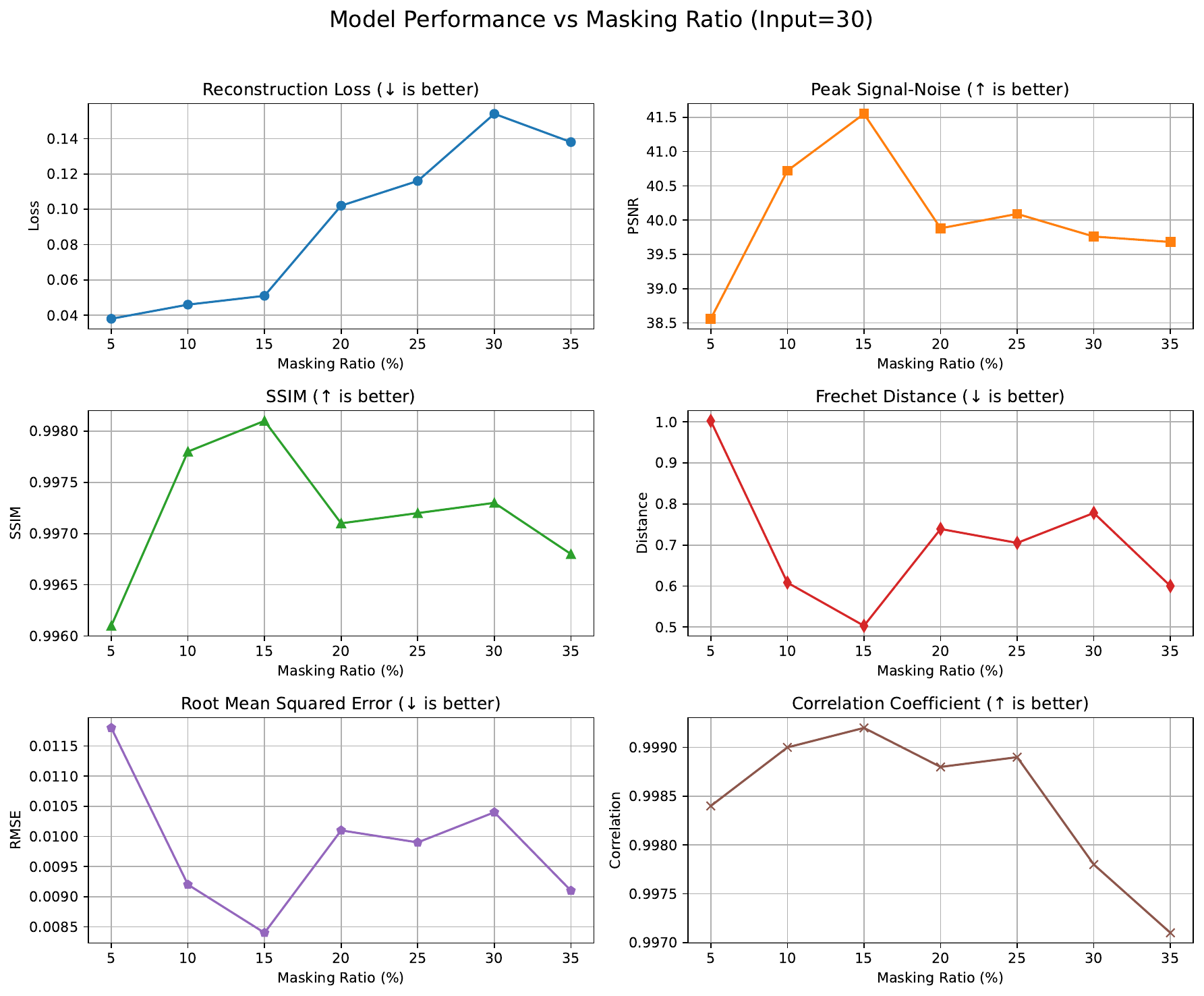}
    \caption{Model performance trends across varying masking ratios (x-axis) with a fixed input size of 30. The six subplots display Reconstruction Loss (RL), Peak Signal-to-Noise Ratio (PSNR), SSIM, Fréchet Inception Distance (FID), Root Mean Squared Error (RMSE), and Correlation Coefficient on the y-axes. Each subplot illustrates how the metric changes with masking ratio ($\gamma$), highlighting the trade-offs in model accuracy and quality.}
    \label{fig:masking}
\end{figure*}
\begin{figure*}[ht!]
    \centering
    \includegraphics[width=0.98\linewidth]{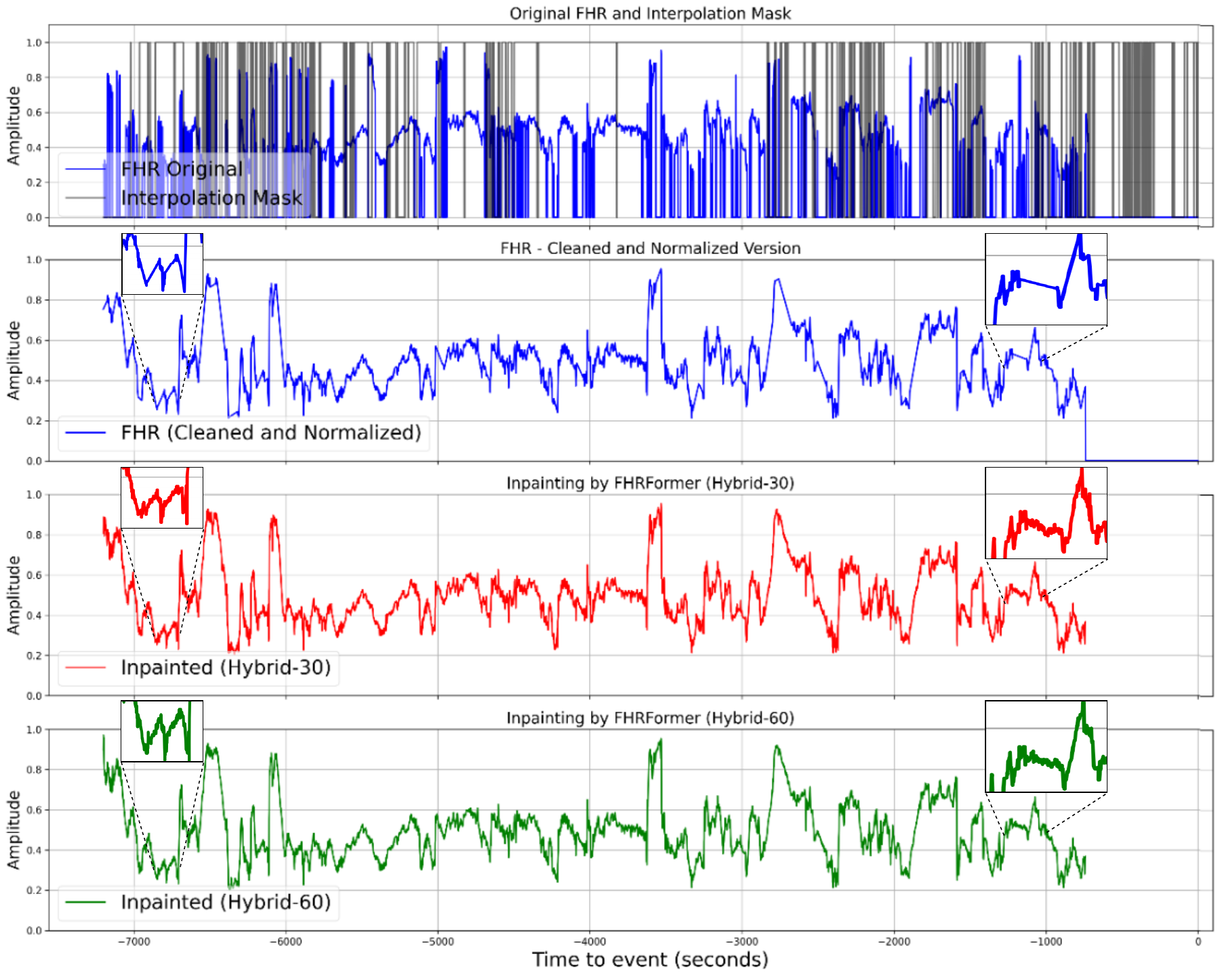}
    \caption{Inpainting application using the  FHRFormer. The first row shows the original FHR signal from the Moyo device. Vertical black lines indicate samples removed during pre-processing (denoising process). The second row shows the noise-removed and interpolated version. The third and fourth rows show inpainting performance by the best (Hybrid-30) and the second-best (Hybrid-60).}
    \label{fig:inpainting}
\end{figure*}

\begin{figure*}[h!]
    \centering
    \includegraphics[width=0.98\linewidth]{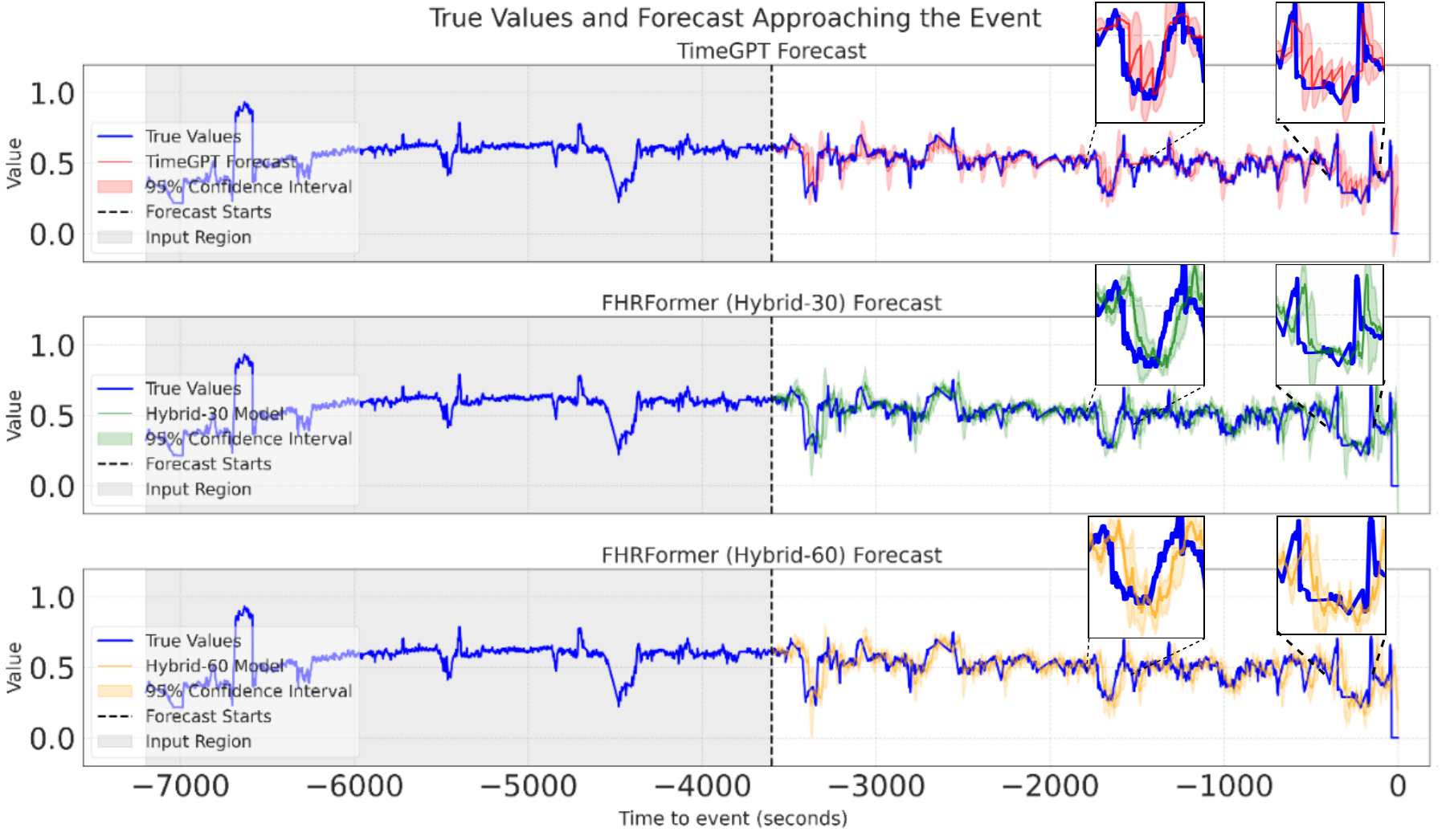}
    \caption{Forecasting application using the FHRFormer. The FHRFormer is fed with a context window (gray region), which is 3600 timesteps (30 minutes of FHR data) as the past horizon, and starts forecasting 30 timesteps (15 seconds of FHR data) in a progressive forecasting style. The first row shows forecasting performance by the TimeGPT~\citep{garza2023timegpt} model. The second row shows forecasting performance by the best FHRFormer, and the last row shows forecasting performance by the second-best FHRFormer.}
    \label{fig:forecasting}
\end{figure*}

\textit{Root Mean Squared Error (RMSE)} provides an interpretable measure by taking the square root of MSE, i.e., $ \mathrm{RMSE} = \sqrt{\mathrm{MSE}} $, allowing direct comparison on the scale of the input values. \textit{Mean Absolute Error (MAE)} reflects the average magnitude of reconstruction errors without considering their direction: $ \mathrm{MAE} = \frac{1}{N} \sum_{i=1}^N |\mathbf{x}_i - \hat{\mathbf{x}}_i| $. The \textit{Peak Signal-to-Noise Ratio (PSNR)} expresses the quality of the reconstruction as the ratio between the maximum possible signal and the error power, measured in decibels (dB): $ \mathrm{PSNR} = 10 \cdot \log_{10}\left(\frac{{\max(x)^2}}{\mathrm{MSE}}\right) $. Higher PSNR values indicate better reconstruction fidelity. The \textit{Structural Similarity Index Measure (SSIM)} evaluates perceived similarity between original and reconstructed signals considering luminance, contrast, and structure, with SSIM values ranging from 0 to 1, where 1 represents perfect similarity.

\textit{The Fréchet Inception Distance (FID)} quantifies the distributional similarity between the original and reconstructed feature representations. Lower FID scores correspond to higher fidelity in the reconstructed data distributions. The \textit{Correlation Coefficient (CC)} measures the linear correlation between ground truth and reconstructed signals as $ \mathrm{CC} = \frac{\mathrm{cov}(\mathbf{x}, \hat{\mathbf{x}})}{\sigma_x \sigma_{\hat{x}}} $, where $\mathrm{cov}$ denotes covariance, and $\sigma_x$, $\sigma_{\hat{x}}$ are the standard deviations of the true and reconstructed signals, respectively.

To ensure the integrity of the performance metrics, the MSE, RMSE, and MAE are calculated exclusively from the originally observed data points held out during the masking process. Any pre-processed linear interpolations used for dimensionality consistency are excluded from the final error calculation to prevent bias towards the interpolation method. Moreover, before computing evaluation metrics, reconstructions were rescaled back to the original range using the same per-signal min-max values, ensuring MSE, RMSE, and MAE are reported in the original units (bpm). All metrics are reported using model weights corresponding to the lowest validation loss observed during training. No hyperparameter tuning was performed on the held-out test set.

% ------ 

\section{Results} 
\label{sec:results}

In this section, we present the empirical results of the self-supervised training phase of FHRFormer. The optimized model parameters are evaluated across two downstream evaluation tasks: i) Inpainting, which targets the reconstruction of missing FHR signal segments, and ii) Forecasting, which evaluates progressive prediction capabilities over future horizons. 

\subsection{Self-supervised FHR Encoder Configured via Masking}
The reconstruction performance of FHRFormer was evaluated across multiple input patch configurations to isolate the structural scale required for signal tracking. Table~\ref{tab1} presents an empirical comparison of eight performance metrics across experimental patch sizes ($p_s$) ranging from 30 to 480 samples.

The validation metrics show an inverse relationship between patch size and reconstruction accuracy across both splits. The configuration utilizing $p_s=30$ recorded the lowest error profiles and highest cross-correlation values. Based on these data benchmarks, the model trained with a patch size of 30 is designated as \textbf{\textit{Hybrid-30}}, while the second-best configuration utilizing $p_s=60$ is designated as \textbf{\textit{Hybrid-60}}.

Figure~\ref{fig:masking} tracks the model metrics across six evaluation criteria as a function of the target masking ratio ($\gamma$). The reconstruction profile reaches its performance peak when the masking ratio is configured at $\gamma=15\%$. Performance degrades continuously as the masking ratio is increased beyond this optimal operational threshold.

\subsection{Inpainting Task Performance}
The pre-trained FHRFormer architecture was deployed to resolve missing data intervals within the pre-processing pipeline. The sequence boundaries matching missing data dropouts were treated as masked indices, allowing the model to reconstruct full-length continuous FHR sequences where values are updated exclusively at the masked coordinates.

Figure~\ref{fig:inpainting} illustrates the comparative inpainting performance on a sample recording. The first row displays the raw FHR signal obtained via the Moyo device along with a binary mask mapping the dropped data regions. The second row displays the baseline linear interpolation processing step. The third and fourth rows show the inpainted signals generated by the Hybrid-30 and Hybrid-60 models, documenting the variations reconstructed across the missing data boundaries.

\subsection{Forecasting Task Performance}
To assess predictive sequence mapping, the trained latent weights of FHRFormer were coupled to a recursive progressive forecasting module. The architecture processes a fixed historical sequence window to project future states, where each estimated block is iteratively appended to the input sequence vector to generate successive short-term predictions. Our hyperparameter tuning established a 30-minute historical context window (3600 timesteps) combined with a 15-second predictive horizon (30 timesteps) as the evaluation baseline. This short predictive horizon reflects FHRFormer's intended deployment as an embedded alarm feature within the Moyo device, where forecasts must flag an emerging deceleration or instability within seconds to be clinically actionable, rather than support long-range trend prediction. During development, we evaluated candidate horizons and observed that shorter windows produced more reliable, lower-uncertainty forecasts suited to this alarm-triggering use case, which motivated adopting the 15-second horizon reported here.

Figure~\ref{fig:forecasting} provides a visual tracking breakdown of a test-set FHR sequence compared against the TimeGPT base model configuration. The historical context is bounded by the shaded gray region. The blue curve tracks the true physiological signal trajectory, whereas the color-coded paths track the model projections. The outer shaded boundaries display the empirical error bounds derived from the standard deviation computed across the validation set residuals over the 15-second forecasting horizon.

\section{Discussion}  \label{sec:discussion}
The empirical behavior of FHRFormer across varying hyperparameter configurations demonstrates that signal reconstruction accuracy is governed by a fundamental trade-off between patch size and macro-structural contextual modeling. By establishing that a sequence patch size of 30 samples yields optimal fidelity, the experimental data reveal a critical operational threshold where the model balances local morphological resolution against sequence trends. When patches are expanded toward larger dimensions, the model loses its capacity to capture the rapid, transient oscillations characteristic of true FHR signals. This optimization directly explains the performance trends observed with the \emph{Hybrid-30} and \emph{Hybrid-60} models compared to traditional linear interpolation baselines. Clinically, such flattening is highly problematic because micro-variability serves as an essential indicator of fetal health status. Because the multi-head self-attention layers compute temporal dependencies across the full sequence window, the architecture utilizes unmasked anchors to infer missing physiological patterns, offering a robust mechanism to counter the short-to-medium signal dropouts caused by maternal movement or sensor displacement.

The evaluation across different masking ratios indicates that the performance peak at 15\% represents an optimal balance for the self-supervised reconstruction objective. When larger portions of the FHR signal are withheld, the missing data density begins to degrade the structural context available to the encoder, hindering its capacity to compute precise latent mappings. This finding aligns well with the practical constraints of wearable monitoring systems, where dropouts typically occur as localized, transient blocks rather than total signal loss. A primary technical driver of this morphological consistency is the implementation of a dual-domain objective framework that bridges time-domain and frequency-domain constraints. Standard temporal loss functions often force a predictive model to optimize for global amplitude alignment, an optimization bias that remains indifferent to phase shifts and spectral distribution. Incorporating the focal frequency loss directly penalizes spectral deviations in the discrete Fourier domain, compelling the decoder to map the true power distribution of the original signal and yielding superior scores across spectral metrics such as PSNR, SSIM, and FID.

The adaptation of the pre-trained attention weights to progressive forecasting demonstrates that the model transcends trivial temporal persistence loops, showing a sophisticated capacity to model forward-looking physiological trajectories. By parsing a fixed 30-minute historical context window to iteratively project a 15-second future horizon, the model leverages its learned internal representation of fetal heart rate dynamics to track complex variations rather than defaulting to steady-state assumptions. The stability of this recursive prediction loop, where each newly estimated segment is sequentially fed back into the context sequence to update successive attention states, indicates that the latent feature space is resilient against rapid error accumulation and catastrophic drift over short-term lead times. This temporal agility allows the model to outperform generalized time-series foundation architectures such as TimeGPT, which struggle to adapt to the highly non-stationary accelerations and decelerations unique to intrapartum physiology. In a real-time deployment context, this short-term predictive capability introduces a vital buffer, allowing wearable devices to sustain continuous automated risk assessments or anticipate decelerations during transient blackouts in data acquisition. Consistent with this deployment objective, we deliberately constrained FHRFormer's evaluation to short, clinically actionable forecasting windows rather than arbitrarily long horizons: a wearable alarm system gains little from a forecast that is only reliable minutes ahead, and the compounding uncertainty inherent to recursive forecasting makes shorter horizons the more clinically appropriate operating point for this application.

\textit{Limitations:} The current model reconstructs short-to-medium gaps in FHR signals recorded with the Moyo device. A limitation of the architecture is that it does not provide probabilistic uncertainty estimates. Future iterations should integrate generative or Bayesian layers to offer explicit confidence intervals alongside the reconstructed traces. In addition, reconstruction accuracy was evaluated on artificially masked segments of observed signal, since true values are not available for real dropouts. Finally, all recordings were collected with Moyo monitors in Tanzanian hospitals, and performance on other monitor types and populations remains to be established.

\textit{Clinical relevance}: The long-term clinical aims of this work are to identify during labour when intervention, such as caesarean section, is needed, and to predict which newborns will need help to breathe at birth. Wearable monitors such as Moyo make it feasible to collect continuous FHR recordings from thousands of labours, also in low-resource settings. However, these recordings often contain gaps when the sensor is displaced as the mother moves. A simple workaround, linear interpolation, replaces missing segments with straight lines that remove beat-to-beat variability, a feature clinicians rely on when judging fetal wellbeing. FHRFormer may contribute in two ways. Its immediate contribution is to retrospective research. By filling gaps while better preserving features such as variability and deceleration shape, it allows incomplete recordings to be used when developing models that predict the need for intervention or resuscitation. In the longer term, the same approach could run within a monitor during labour to provide more complete traces in real time, supporting clinicians or automated risk assessment at the bedside. We have not yet tested whether clinicians interpret reconstructed segments as reliably as recorded ones. The forecasting results indicate that the model has learned meaningful FHR dynamics. Whether short-term forecasting could support clinical monitoring, for example by bridging brief signal dropouts, remains to be investigated. This study does not evaluate clinical decisions or neonatal outcomes, and any clinical use would require prospective validation.

\section{Conclusion and Future Work}  \label{sec:conclusion}
In this paper, our contributions are threefold. 

First, we introduced a transformer-based self-supervised architecture for FHR encoding and reconstruction, the {\bf FHRFormer}, which incorporates a frequency-aware loss function to preserve fidelity in both physiological morphology and spectral domains. This capability is particularly important for reconstructing signal gaps with variable durations. The model was trained on a large dataset comprising 4,486 FHR signals collected using the Moyo device, using a self-supervised masking strategy.

Second, we applied FHRFormer to inpaint real missing segments in FHR signals. The reconstructed segments preserve both local temporal and spectral characteristics consistent with the surrounding signal, improving continuity and quality for downstream analysis.

Third, we proposed a forecasting approach using FHRFormer to predict future FHR values. This strategy enables proactive signal assessment and has the potential for real-time monitoring and early warning in clinical settings.

Future work includes validation of the proposed methods in clinical settings and potential extension to a multimodal framework that incorporates additional clinical parameters or maternal measurements.  

\section*{Ethical Approval}
The Safer Births 2  project was ethically approved before implementation by the National Institute of Medical Research (NIMR) in Tanzania (NIMR/HQ/R.8a/Vol. IX/3852) and the Regional Committee for Medical and Health Research Ethics (REK) in Norway, reference 172126. 

Conflict of Interest: K.E. has a 20\% affiliation with Laerdal Medical; however, the research was conducted independently of this affiliation. The remaining authors declare that they have no conflicts of interest.

\section*{Acknowledgment}
The Laerdal Foundation and the Research Council of Norway, through the Global Health and Vaccination Program (GLOBVAC - project number 228203), funded the research infrastructure, data collection, and management of the Safer Births 2 research project. The Idella Foundation funded the postdoctoral grant of N.K. 
The authors also thank all mothers and midwives who contributed to the data collection.
ChatGPT and Writefull have been used for minor editing and grammar improvements.  All intellectual content and interpretations are solely those of the authors.

\bibliographystyle{plainnat}
\bibliography{main}
\end{document}